\documentclass{journal}

\usepackage[a4paper, total={6in, 8in}]{geometry}

\usepackage{url,hyperref,lineno,microtype,subcaption}
\usepackage[onehalfspacing]{setspace}
\usepackage[english]{babel} 
\usepackage{algorithm} 
\usepackage{comment}
\usepackage[algo2e]{algorithm2e} 
\usepackage[warn]{textcomp}

\usepackage{graphicx}
\usepackage{inputenc}
\usepackage{crop}\usepackage{amsmath}
\usepackage{array}
\usepackage{color}
\usepackage{amssymb}
\usepackage{flushend}
\usepackage{stfloats}
\usepackage{amsthm}
\usepackage{chngpage}
\usepackage{times}
\usepackage{datetime}
\usepackage{parskip}
\usepackage{epstopdf}
\usepackage{authblk}
\usepackage[numbers]{natbib}

\usepackage[normalem]{ulem}

\newcommand{\kmeans}{{\sc k-means}\xspace}
\newcommand{\nkm}{{\sc N_{km}}\xspace}
\newcommand{\Msm}{{\sc \mathbf{\Lambda_{sm}}}\xspace}
\newcommand{\Ms}{{\sc \mathbf{\Lambda_{s}}}\xspace}
\newcommand{\Tsm}{{\sc \mathbf{T}}\xspace}
\newcommand{\pmin}{{\sc p_{min}}\xspace}

\newcommand{\Nmv}{{\sc N_{mv}}\xspace}
\newcommand{\nspec}{{\sc N}\xspace}
\DeclareMathOperator{\cut}{cut}
\DeclareMathOperator{\Ncut}{Ncut}
\DeclareMathOperator*{\argmax}{argmax}

\begin{document}

\title{Identification of Invariant Sensorimotor Structures as a Prerequisite for the Discovery of Objects}

\author[1]{Nicolas Le Hir\thanks{nicolas.lehir@softbankrorobtics.com}}
\author[2]{Olivier Sigaud\thanks{Olivier.Sigaud@upmc.fr}}
\author[1]{Alban Laflaqui\`ere\thanks{alaflaquiere@softbankrobotics.com}}
\affil[1]{AI Lab, SoftBank Robotics Europe, Paris, France}
\affil[2]{Sorbonne Universit\'e, Institut des Systèmes Intelligents et de Robotique, CNRS UMR 7222, Paris, France}
\renewcommand\Authands{ and }

\maketitle

\begin{abstract}
Perceiving the surrounding environment in terms of objects is useful for any general purpose intelligent agent. In this paper, we investigate a fundamental mechanism making object perception possible, namely the identification of spatio-temporally invariant structures in the sensorimotor experience of an agent. We take inspiration from the Sensorimotor Contingencies Theory to define a computational model of this mechanism through a sensorimotor, unsupervised and predictive approach. Our model is based on processing the unsupervised interaction of an artificial agent with its environment. 
We show how spatio-temporally invariant structures in the environment induce regularities in the sensorimotor experience of an agent, and how this agent, while building a predictive model of its sensorimotor experience, can capture them as densely connected subgraphs in a graph of sensory states connected by motor commands.
Our approach is focused on elementary mechanisms, and is illustrated with a set of simple experiments in which an agent interacts with an environment. We show how the agent can build an internal model of moving but spatio-temporally invariant structures by performing a Spectral Clustering of the graph modeling its overall sensorimotor experiences.
We systematically examine properties of the model, shedding light more globally on the specificities of the paradigm with respect to methods based on the supervised processing of collections of static images.
\end{abstract}

\section{Introduction}
\label{sec:Introduction}

Humans flexibly interpret their rich sensorimotor experience of the world in terms of objects in the environment. In that respect, we assume that this ability to discover, identify, and manipulate objects is required for any general purpose intelligent robot.
Despite great progress in object detection \citep{Redmon2015} or classification \citep{He2016} in the last few years, the computer vision community still lacks a clear formalization of the problem of autonomous object identification by an artificial agent.
Understanding the fundamental nature of objects and their perception is a core philosophical question that we do not pretend to fully address in this work. Rather, we focus on a specific property that we assume plays an important role in the above question: the spatio-temporal invariance of objects. More precisely, we propose to investigate a mechanism assumed to be fundamental for autonomous object perception, namely the unsupervised identification of invariant spatio-temporal structures in the sensorimotor flow of an agent.

Perception, and in particular artificial perception, is traditionally considered as a passive process in which the sensory state obtained through sensors is projected onto higher-level representations, which in turn inform higher-level cognitive processes which generate actions. This perspective has however been challenged by multiple philosophers and neuroscientists who claim that perceptive experience emerges from internal predictive modeling of the sensorimotor interaction with the environment \citep{Vonhelmholtz1896, Gibson1979, Friston2006, Clark2013}
Our work fits in with such a predictive and sensorimotor description of perception. It is based on two prominent theories, namely the Sensorimotor Contingencies Theory (SMCT) \citep{ORegan2001} and Predictive Coding \citep{Rao1999, Rao2005}. The former claims that perception is based not only on sensory information but also on the knowledge of regularities in the way an agent's actions can transform its sensory inputs. The latter suggests that the brain hierarchically builds a predictive model of the causes of its sensory experience. The two viewpoints align nicely when considering that regularities in the sensorimotor flow can be used as support for a predictive model \citep{Seth2014}.  
\\
In this framework, we focus on an elementary property of objects and we study how this property can be exploited to contribute to their discovery by extracting regularities in the sensorimotor experience of an artificial agent. Namely, we assume that objects have an intrinsic structure which is spatio-temporally invariant, and limited in space. In that respect, we assume on the one hand that the intrinsic properties of objects, such as shape, size, or appearance, are preserved across time and space. On the other hand, being limited in space simply means that the objects are smaller than the world explored by the agent.

This spatio-temporal stability of objects implies structure in the sensorimotor experience an agent has when interacting with them. This way, observing one part of a known object, the agent can predict what would be observed on other parts of this object. For example, seeing one side of a tomato, it can predict what the other side of the tomato would look like, as put forward through the concept of \emph{perceptual presence} in \citep{Seth2014}. According to the SMCT, this property is constitutive of the experience of objects \citep{Oregan2001a}.

In this paper, we propose a minimalistic simulation in which an agent visually explores  in a random way an environment containing spatio-temporally invariant structures. We assume having a spatio-temporally invariant structure is one generic property of objects, but it may not be the only one. Hence we refer to identifying these spatio-temporally invariant structures as identifying \emph{proto-objects} in the rest of the paper. 
Admittedly, as long as the decisions of our agent are random and its actions only consist of visual exploration, we may consider our work from the perspective of pattern identification in signal processing (see e.g. \citep{Jain2000}). However, we present this work from an agent-based perspective for three reasons.  First, in our framework, an agent is generically “that which acts”: it is sufficient that it produces actions to be considered as an agent. Second, in Section~\ref{sec:rotating}, we investigate a case where the agent actively rotates objects in its environment. Third, the case of an agent deciding which future action is optimal according to a goal is an important step in our future work agenda.

In our simulations, the world explored by the agent can change in two ways. First, the proto-objects, while keeping their internal structure, can move randomly in the world, or even be introduced/removed. Second, the rest of the environment can itself change randomly.
Importantly, despite these changes, the world is statistically invariant enough so that the agent is able to partially explore it between two successive changes.
This setup, illustrated in \figurename{~\ref{fig:explo}}, can intuitively be interpreted as having proto-objects that can move in the environment, and can be encountered in different contexts.
Our model is minimalistic in the sense that we assume no prior knowledge on the world or on the agent itself, neither on its spatial structure, on the environment structure, nor on the proto-objects. The naive agent follows a random exploration policy, and interacts in a generic way with an external environment through an interface of uninterpreted sensorimotor information \citep{Hoffman2016, Rafael2017}.
In line with Predictive Coding, we propose a method for the agent to build a sensorimotor predictive model of its exploratory experience, and to identify the sensorimotor regularities induced by the proto-objects. 
More precisely, we model the sensorimotor experience as a weighted multigraph in which the nodes correspond to sensory states, and each pair of states is linked by several edges representing different motor commands. The weight of each edge corresponds to the conditional probability of the corresponding sensorimotor transition. Regularities in the sensorimotor interaction with the environment should then appear as stronger connections between some pairs of nodes. In particular, we hypothesize that the presence of proto-objects should induce the presence of some densely intra-connected subgraphs that the agent can identify as its own experience of these proto-objects. The representation of these regularities can then be used by the agent for counterfactual prediction, which makes the identification of proto-object a worthy objective.

The paper is organized as follows. In Section~\ref{sec:Methods}, we describe a simple simulation to illustrate the approach, as well as a computational method to identify the sensorimotor regularities induced by proto-objects. 
In Section~\ref{sec:res}, the results produced by the method applied to the simulated system are thoroughly presented. Additional experiments are also designed to highlight the properties and limitations of the approach. Finally, in Section~\ref{sec:Discussion}, we discuss the benefit of our paradigm with regards to the perception of objects. We also consider the future steps that would extend the current illustrative simulation towards more complex and realistic setups.
This work is a direct extension to the preliminary results presented in \citep{Laflaquiere2015a, Hemion2017}.

\section{Method}
\label{sec:Methods}

In this section, we introduce a simplistic simulation in which an agent explores an environment containing proto-objects. We then propose a method to process its sensorimotor experience and identify the regularities induced by these structures.

\subsection{Simulation}
\label{sub:Simulation}

\begin{figure}[t!]
    \centering
    \includegraphics[width=1\textwidth]{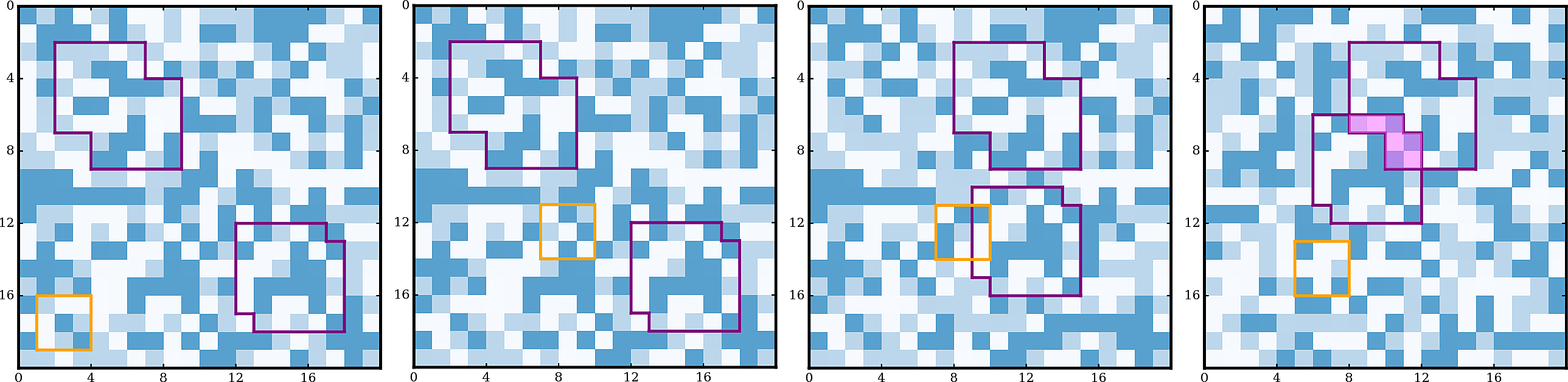}
    \caption[Explorations]{\textbf{Simulation setup and four example consecutive exploration steps:} The position of the agent’s sensor is outlined in orange, whereas the proto-objects are outlined in purple. The agent’s sensor moves at each time step (but since the null movement is possible, it has a non-zero probability of keeping the same position). At each time step, each proto-object has a probability to move, independently from the other proto-objects and the rest of the environment. At each time step, the rest of the environment has a probability to randomly change.
	Here for instance, the agent moves at steps 2 and 4. At step 3 both proto-objects move, and at step 4, only one of the them moves, partially overlapping the other proto-object (zone highlighted in purple). At step 4 the environment also changes. Note that the purple outline of the objects is added here for visualization and that the agent does not have any access to it.}
    \label{fig:explo}
\end{figure}

The simulation we propose  consists in an agent exploring an environment containing proto-objects . The environment is a two-dimensional square gridworld of fixed size $20\times 20$ discrete elements, or "pixels". Each pixel can takevalues in $\{1,2,3\}$, and is initialized randomly at the beginning of the simulation.
At each time step, the environment can change with a probability $p_{env}=0.05$, in which case the values of all its pixels are randomly redrawn. At the beginning of the simulation, $N_{obj}=2$ proto-objects are created in the environment. They correspond to $N_{obj}$ sets of contiguous pixels drawn from the same distribution as pixels of the environment, but which keep the same internal structure during the whole simulation. They are of minimum size $5\times 5$ and maximum size $7\times 7$, and do not necessarily have a square shape, as illustrated in \figurename{~\ref{fig:explo}}.
During the simulation, the proto-objects are moved in the environment with a probability $p_{obj}=0.1$ at each time step. Furthermore, they can be independently removed from the environment with a probability $p_{abs}=0.2$.
If present, the proto-objects pixels occlude those of the environment, and also potentially occlude each other, as illustrated in \figurename{~\ref{fig:explo}}.
{Note that an agent cannot distinguish proto-objects from the environment simply based on a single sensory input, since the pixels that constitute them are drawn from the same distribution. They only differ in the spatio-temporal consistency that proto-objects maintain in contrast with the environment during the simulation.
}\\
The agent observes this two-dimensional world with a  limited sensor, which is a $3\times 3$ patch window, through which it receives sensory inputs. It can move its sensor anywhere in the environment, using motor commands. At each time step, the sensorimotor input of the agent contains a sensory input $\mathbf{s}_t$ (which is a 9-dimensional vector of pixels) and a motor command $\mathbf{m}_t$ (which is a 2-dimensional vector, representing the horizontal and vertical components of the sensor displacement in the visual scene). Together with the sensory input $\mathbf{s}_{t+1}$ experienced after performing $\mathbf{m}_t$, these sensorimotor experiences form this sensorimotor experience forms a \emph{sensorimotor transition triplets triplet} $(\mathbf{s}_t,\mathbf{m}_t,\mathbf{s}_{t+1})$. 

At each time step of the simulation, the agent moves its sensor by randomly picking a new position in the environment (possibly the same as the current one), and stores the experienced sensorimotor triplet. Since the environment and the proto-objects change with a lower probability than the agent, the latter can sensor position, the agent can statistically explore their content over several time steps and extract the regularities they induce.

\subsection{Processing method}
\label{sub:processing}

We now describe the way the agent processes its sensorimotor experience in order to identify proto-objects in the environment. First, the data are compacted by a clustering step. Then, the sensorimotor transitions are stored in a three-dimensional tensor, representing a statistical model of the agent's sensorimotor experience.  This tensor is analyzed to extract densely connected subgraphs.

\subsubsection{Storing of the sensorimotor experience}

\begin{figure}[t!]
    \centering
    \includegraphics[width=1\textwidth]{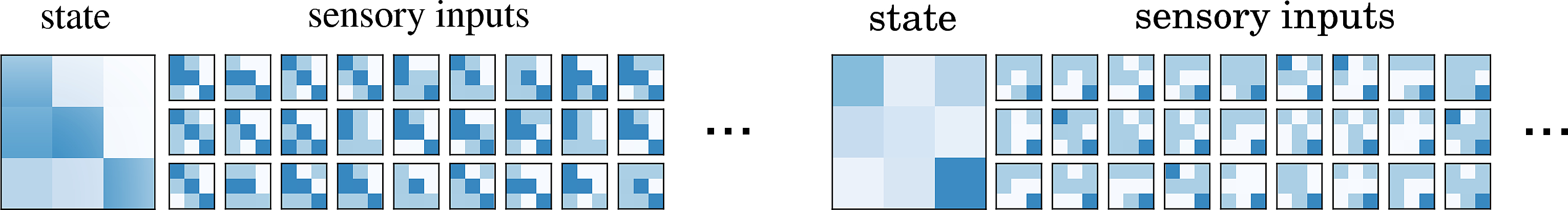}
    \caption[Two examples of \kmeans]{\textbf{Examples of k-means clustering:} Two states and example sensory inputs associated to each of these states by k-means clustering.}
    \label{fig:kmeans}
\end{figure}

The agent interacts with the environment during $n_{step}=3e7$ steps and its sensorimotor experience is stored and processed off-line.
We store the empirical conditional probabilities $p(\mathbf{s}_{t+1}|\mathbf{s}_t,\mathbf{m}_t)$ of each sensorimotor triplet $(\mathbf{s}_t,\mathbf{m}_t, \mathbf{s}_{t+1})$ experienced by the agent in a three-dimensional tensor $\Tsm$. In $\Tsm$, $\mathbf{s}_t$ and $\mathbf{s}_{t+1}$ correspond to the row and the column respectively, while $m$ is a one-dimensional encoding of the movement performed at time $t$ and corresponds to the depth in the tensor. However, in order to limit the size of $\Tsm$ and the computational cost of the simulation, the representation of the sensory experience is compacted beforehand by clustering together similar sensations. We use a simple \kmeans algorithm to perform this clustering, where the number of clusters is arbitrarily set to $\nkm=250$. These clusters group together the sensory inputs considered by the agent to build its predictive model, as illustrated in \figurename~\ref{fig:kmeans}. In the following, the resulting centroids produced by the \kmeans clustering algorithm are called "states". The number of possible movements the agent can perform in this $20\times 20$ environment is $\Nmv=1024$. Thus, the size of the tensor $\Tsm$ is $(\nkm \times \nkm \times \Nmv)=(250\times 250 \times 1024)$. 

\subsubsection{Densely connected subgraph identification}
\label{sec:simi}

The tensor $\Tsm$ can be seen as an approximation of the weighted graph mentioned in Section \ref{sec:Introduction}, in which the weights of the multiple edges between two nodes are the conditional probabilities of the corresponding transition, labeled by the action. We want to identify densely connected subgraphs of sensory states in this graph.
To do so, we propose to use Spectral Clustering \citep{Planck2006, Meila2015}, which requires the definition of a similarity between each pair of nodes $(\mathbf{s}_a,\mathbf{s}_b)$ in the graph.\\
Intuitively, two nodes will be considered similar if a transition between them is experienced with a high enough probability. In order to define the similarity, we first filter out some transitions lacking statistical relevance, by discarding rows $T[\mathbf{s}_a,:,\mathbf{m}]$ such that 
the movement $\mathbf{m}$  has been performed less than $n_{min}=20$ times while experiencing state $\mathbf{s}_a$. Then, for other triplets, let $\mathcal{E}$ be the subset of sensorimotor transitions: $\mathcal{E}=\{(\mathbf{s}_a,\mathbf{m},\mathbf{s}_b)\;|\;p(\mathbf{s}_b|\mathbf{s}_a,\mathbf{m}) \;\geq\;~p_{sim}\}$, where $p_{sim}$ is a threshold  set to $0.3$.
We define the sensorimotor similarity $\Msm(\mathbf{s}_a,\mathbf{s}_b)$ between each pair of states $(\mathbf{s}_a,\mathbf{s}_b)$ as:

$$\Msm(\mathbf{s}_a,\mathbf{s}_b)=\sum_{m \in \mathcal{E}}p(\mathbf{s}_b|\mathbf{s}_a,\mathbf{m}).$$

\noindent
Applying this method to all pairs of states, we derive the 2D sensorimotor similarity matrix $\Msm$. Finally, since similarities are usually defined for undirected graphs, we make $\Msm$ symmetric by averaging it with its transpose. This procedure is formally summarized in Algorithm~\ref{al:algo}. We then apply Spectral Clustering to the graph defined by the similarity $\Msm$. Spectral Clustering is a graph clustering method that is often used when the relation between the nodes of the graph is quantified by a general measure of similarity, that is not necessary a distance. To define the clusters, the eigenvectors of the Laplacian of the graph are computed. A change of representation is then performed by building new vectors with the components of the main eigenvectors of the Laplacian. A regular clustering is then performed in the space corresponding to these new vectors, yielding the final clusters. More details can be found in \mbox{
\citep{Planck2006, Meila2015}}
.

\begin{algorithm}[t!]
\KwData{$3D$ tensor $\Tsm$ of sensorimotor transitions}

 initialize empty matrices $\Msm$ and $\Ms$ of size $\nkm \times \nkm$ all entries set to $0$\;

\For{$\mathbf{s}_a$ in $1..\nkm$}{
     \For{$\mathbf{m}$ in $1..N_{mv}$}{
        $row=\Tsm[\mathbf{s}_a,:,\mathbf{m}]$\;

        $\Ms[\mathbf{s}_a,:]=\Ms[\mathbf{s}_a,:]+row$\;

         \If{$sum(row)>n_{min}$}{
             $p_{max}=max(row)$\;

             $\mathbf{s}_{b_{max}}=\argmax (row)$\;

             \If{$p_{max}>\pmin$}{
                 $\Msm[\mathbf{s}_a,\mathbf{s}_{b_{max}}]=\Msm[\mathbf{s}_a,\mathbf{s}_{b_{max}}]+p_{max}$
             }
         }
     }
 }

 $\mathbf{\Msm}=\frac{1}{2}(\mathbf{\Msm}+\mathbf{\Msm}^T)$

 $\mathbf{\Ms}=\frac{1}{2}(\mathbf{\Ms}+\mathbf{\Ms}^T)$
  \caption{: Building the similarity matrices $\Msm$ and $\Ms$}
 \KwRet{Similarity matrices $\Msm$ and $\Ms$ between sensory states }
 \label{al:algo}
 \end{algorithm}

\subsubsection{Extracting the number of clusters}
\label{sec:clu}

Spectral Clustering requires the specification of the returned number of clusters. Since we wish to introduce as little supervision as possible in our algorithm, we propose to automatically determine it. There is no universal criterion to automatically determine the relevant number of clusters in a general situation, and most criteria are heuristics \citep{Planck2006}. We propose to use the \emph{cut gap} criterion \citep{Meila2015}. The cut gap is identified by finding a knee in the curve of the \emph{normalized cut} as a function of the number of clusters. Consider a graph $G$ clustered in $\nspec$ clusters, forming a clustering denoted $\mathcal{C}=(C_1,\dots,C_{\nspec})$. Given $\mathcal{C}$ and a graph similarity $\Lambda_{ij}$ between each pair of nodes $i$ and $j$, the normalized cut $\Ncut(\nspec)$ is a measure of the quality of $\mathcal{C}$. The lower $\Ncut$, the better the clustering: if $\Ncut$ is very low, it means that the clusters are very weakly connected between each other. It is defined as: 

\begin{equation}
\label{eq:normalized}
\Ncut(\mathcal{C})=\sum_{k=1}^{\nspec}\frac{\cut(C_k,G\backslash C_k)}{d_{C_k}}
=\sum_{k=1}^{\nspec}\frac{\sum_{i\in C_k}\sum_{j\in G\backslash C_k}\Lambda_{ij}}{\sum_{i\in C_k}\sum_{j\in G}\Lambda_{ij}},
\end{equation}
\noindent
where the numerator $\cut(C_k,G\backslash C_k)$ is the \emph{cut} between clusters $C_k$ and $G\backslash C_k$, which is a measure of the strength of the connection between $C_k$ and the rest of the graph. The denominator $d_{C_k}$ is the \emph{degree} of $C_k$, which represents the "weight" of the cluster in the graph. Having low $\cut(C_k,G\backslash C_k)$ terms encourages clusters to be weakly interconnected, while having high $d_{C_k}$ terms favors large clusters, which prevents from yielding trivial isolated outliers as clusters. Thus, the normalized cut leads to a compromise between these two tendencies. In order to find the optimal number of clusters $\nspec^{*}$, we automatically detect the largest $\nspec$ which leads to a low $\Ncut$. To do so, we also compute the second order finite difference of $\Ncut$ as a function of $N$, $$\Delta \Ncut(N)=\Ncut(N+2)+\Ncut(N)-2\Ncut(N+1),$$
 and we take the value $\nspec^*$ that yields the maximum result, that is:
$\nspec^* = \argmax_N \Delta \Ncut(N).$
Thus, the minimal value that can be returned by this criterion is 2.


\subsubsection{Visualizing predictions from the tensor}
\label{sec:predictive}

We can also use $\Tsm$ as a predictive model of the agent's sensorimotor experience. When it receives a certain sensory input, it can use the tensor to try to predict the next sensory input for each possible motor command. More formally, say that the agent experiences state $\mathbf{s}_t$ at time $t$. For each motor command $m$, the agent has learned a conditional probability distribution on the next state, $p(:|\mathbf{s}_t,\mathbf{m})$, and it can use this distribution to make predictions.

However, the visualization of the predictive model is non-trivial, since the predictions of two distinct motor commands may overlap each other, given that the receptive field of the agent is made of several pixels. In order to illustrate some predictions below, we use a mixture of the distributions, in the following way. Let us consider a pixel position $z_{t+1}$ out of the scope of the agent's sensor. Given that the size of this sensor is $3\times 3$ pixels, $9$ motor commands predict the future value of $z_{t+1}$. We manually average the predictions of these $9$ motor commands by computing the weighted average of the states predicted with most certainty by each of these $9$ movements. This process requires some knowledge on the sensorimotor structure of the agent, but it is used for illustration purposes only, and not by the agent itself.

\section{Results}
\label{sec:res}

We now present and analyze the experimental results of the simulations. We then explore alternative setups where the performance of the algorithm is more variable, in order to illustrate the robustness of the approach, but also its limitations.

\subsection{Subgraphs extracted from the predictive model}

\begin{figure}[t!]
    \centering
    \includegraphics[width=1\textwidth]{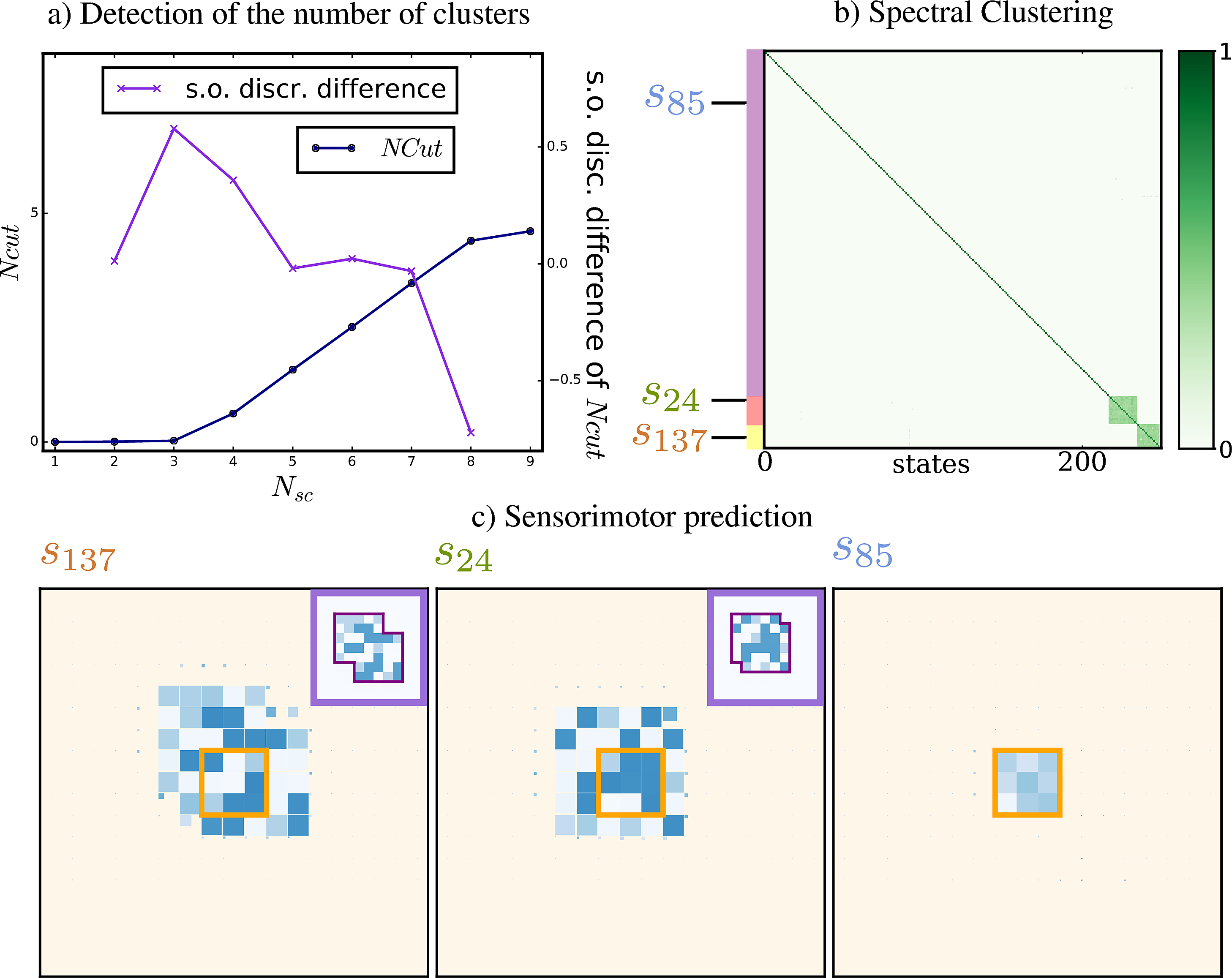}
    \caption[s]
        {\small \textbf{Detection of the number of proto-objects, clustering and prediction:} 
        \textbf{a)} \textbf{Normalized cut and finite difference.} $\Ncut$ (in Black) and second order finite difference of $\Ncut$, $\Delta \Ncut$ (in purple) as a function of the number of spectral clusters. A knee in the $\Ncut$ curve is clearly visible and detected by the second-order derivative at $N^*=3$.
        \textbf{b)} \textbf{Spectral Clustering of the similarity.} The rows and columns of the matrix are reorganized according to the clusters. Three clusters are identified: two densely connected ones corresponding to the proto-objects, and one weakly connected corresponding to the environment. The colored strips at the left of the matrix correspond to the identified identify the different clusters. The similarity scale is represented at the right of the image. The states that are used to visualize the predictive model are indicated by their number.
        \textbf{c)} \textbf{Sensory prediction.} Predictive model of the agent for three input states. If the sensory input is classified as part of a proto-object, the agent can predict its future sensory states as a function of its movement (states $24$ and $137$). If the input state is classified as part of the environment, no probable prediction is made (state $85$).}
    \label{fig:nominal}
\end{figure}

As a reminder, in the simulation, two proto-objects are placed in the environment, with probabilities $p_{obj}=0.1$, $p_{abs}=0.2$, and $p_{env}=0.05$ of movement of the proto-objects, absence of the objects proto-objects, and of change of the environment, respectively. \figurename~\ref{fig:nominal}a) presents the normalized cut $\Ncut$ as a function of $\nspec$, as well as the second order finite difference $\Delta \Ncut(N)$. The curve of $\Ncut$ presents a knee at $\nspec=3$, and the maximal value of the finite difference is attained for  $\nspec^*=3$. Thus, $3$ subgraphs have been identified through Spectral Clustering. This is a good result, since we expect 3 subgraphs to emerge: two subgraphs corresponding to both proto-objects and a third subgraph for corresponding to the environment. \figurename~\ref{fig:nominal} b) shows the similarity matrix, whose lines and columns have been reordered to group together sensory states belonging to the same cluster. The color of each entry in the matrix corresponds to the similarity between both states. Thus, we can visually see that two subgraphs are strongly connected and a third subgraph is weakly connected.

One can also remark that the probabilities on the diagonal of the matrix are higher than elsewhere. This reflects the stability of the world: there is always a non-zero probability that the agent will encounter two identical sensory states consecutively. Let $p(\mathbf{s}_{t+1}=\mathbf{s}_t|m=0)$ be the probability that the agent receives the same sensory input after a time step, given that the agent did not move. If neither of the proto-objects move and the environment did not change, the agent will receive the same input. These conditions being independentbut not simultaneously necessary, we can write that $p(\mathbf{s}_{t+1}=\mathbf{s}_t|m=0)\geq(1-p_{obj})^2(1-p_{env})\simeq 0.76$. This explains the high values found on the diagonal of the similarity matrix.

\subsection{Sensorimotor prediction}

As explained in \ref{sec:predictive}, the three-dimensional tensor built by the agent can be used as a predictive model of its sensorimotor experience. We illustrate it in \figurename{~\ref{fig:nominal}} c), for three input states.
To clarify visualization, the size of each pixel depends on the probability of each prediction : the largest predicted pixels in the figure are the ones predicted with most certainty. In order to compare the prediction with a ground truth, we also show the ground-truth proto-objects introduced in the environment. If the current state was categorized in one of the densely connected clusters, the model successfully reconstructs the total structure of the corresponding proto-object from the small patch it receives: this is the case for instance for states $24$ and $137$. On the contrary, for a state categorized in the third, weakly connected cluster, the model predicts no future sensory state with certainty: this happens for instance for state $85$. 
For such a naïve agent, the objects are thus structures that allow sensorimotor prediction. 


\subsection{Additional experiments}

We propose additional experiments to illustrate the properties and limits of the simulation, the overall approach, and the computational method letting the agent discover proto-objects from its sensorimotor flow. 

\subsubsection{Importance of the motor flow}

\begin{figure}[t!]
    \centering
    \includegraphics[width=1\textwidth]{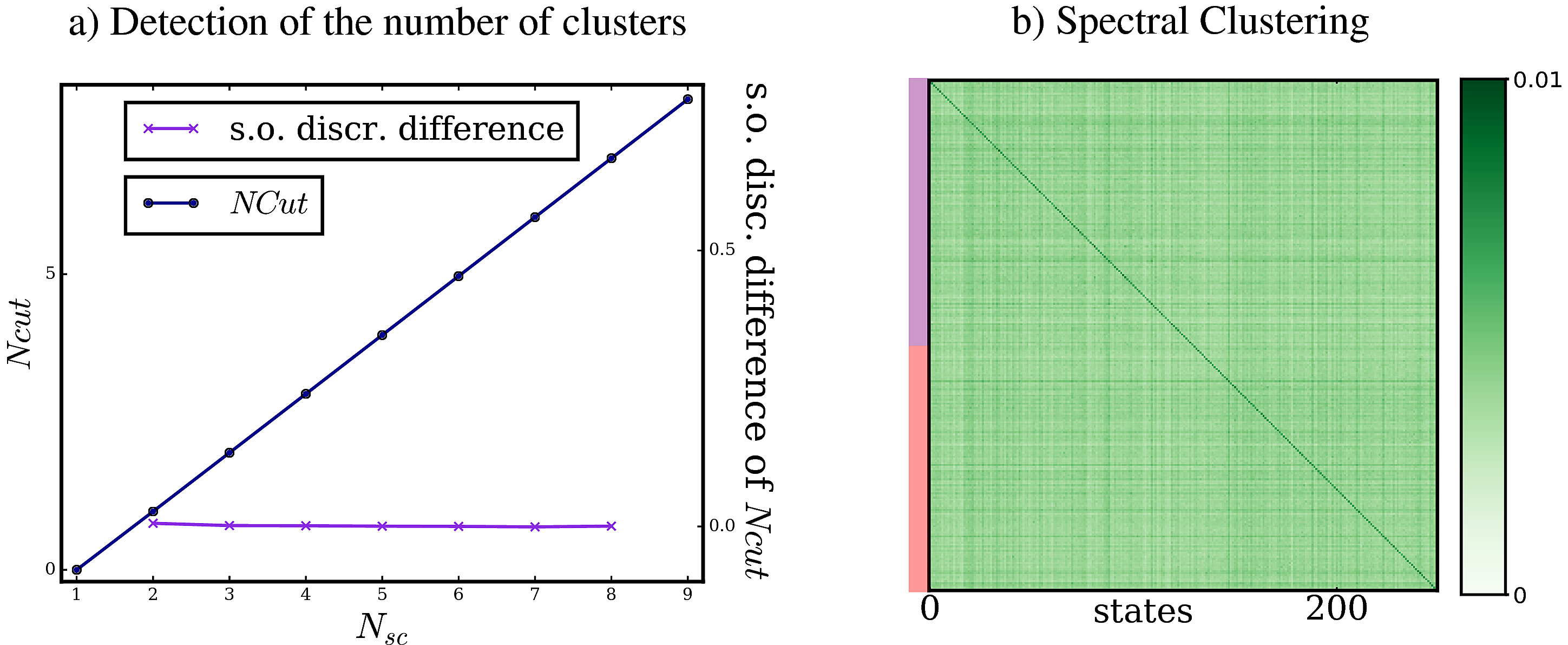}
    \caption[Full matrix prediction]{Detection of the number of proto-objects and Spectral Clustering without motor information:
    \textbf{a)} Applying the normalized cut criterion as presented in \figurename~\ref{fig:nominal} to the sensory similarity $\Ms$ yields poorer results. The number of clusters is not correctly detected by the agent.
    \textbf{b)} Spectral Clustering of the sensory similarity $\Ms$ does not present densely nor weakly connected subgraphs. The states seem to be more uniformly weakly connected (note the change in the colormap scale).
    }
    \label{fig:full}
\end{figure}

In order to illustrate the importance of taking the motor commands into account for discovering proto-objects, we propose a similar processing of the experience of the agent, where motor commands are not recorded by the agent. Instead of the sensorimotor similarity $\Msm$, we derive through Algorithm~\ref{al:algo} a \emph{sensory} similarity: 
$$\Ms(\mathbf{s}_a,\mathbf{s}_b)=p(\mathbf{s}_a\rightarrow \mathbf{s}_b),$$
where $p(\mathbf{s}_a\rightarrow \mathbf{s}_b)$ is the probability of transitioning from state $\mathbf{s}_a$ to $\mathbf{s}_b$, regardless of the motor command. Spectral Clustering of this similarity matrix leads to the results shown in \figurename{~\ref{fig:full}}. The agent is no longer able to detect the correct number of clusters. No clear knee in the cut curve is detected and the cut gap criterion returns two clusters, that is the default outcome of our method when it does not find a cut. In \figurename~\ref{fig:full} b), we see that this "sensory" similarity matrix, even reorganized, does not display any densely connected subgraph. In our setting, the agent is thus unable to extract the structure of proto-objects without using its motor commands.

\subsubsection{Influence of the number of proto-objects}

\begin{figure}[h!]
    \centering
    \includegraphics[width=0.92\textwidth]{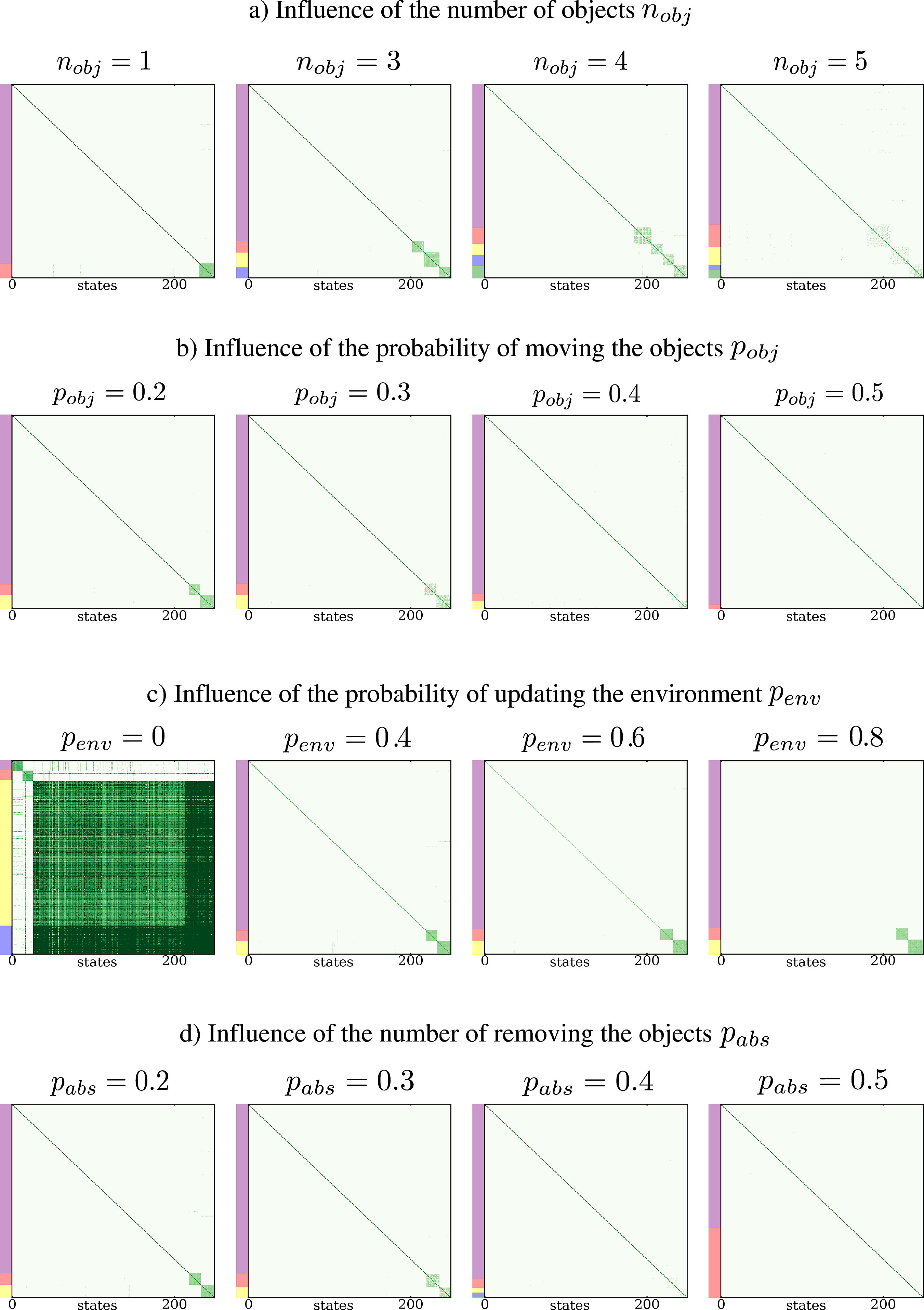}
    \caption[Influence of the simulation parameters]{\textbf{Influence of the parameters of the simulation:} 
    \textbf{a)} The agent is not able to discover and differentiate more than 4 proto-objects in this setup.
    \textbf{b)} When the proto-objects move more frequently, it is harder to extract them from the environment.
    \textbf{c)} The more the environment changes, the easiest it is to extract proto-objects. When the environment never varies, three proto-objects are identified.
   \textbf{d)} When the proto-objects are removed too frequently, their extraction fails and the number of clusters is not correctly determined.
    }
    \label{fig:params}
\end{figure}

We now propose to study the influence of the meta-parameters of the simulation on the results. We first investigate the impact of the number of proto-objects $n_{obj}$ introduced in the environment on the identification of the densely connected subgraphs. The results are shown in \figurename{~\ref{fig:params}} a). For values up to 4, the number of proto-objects is correctly estimated and the clusters are well defined and densely connected. As the number of proto-objects increases, it becomes harder to detect the correct number of proto-objects . If the number is very large, the sensorimotor experience of the agent contains too much randomness and is poorly predictable, since the proto-objects constantly occlude each other in a random order. As a consequence, the probability of consistently experiencing sensorimotor regularities associated with a given proto-objects becomes very low. Here, we see that for $n_{obj}\geq 5$, the Spectral Clustering algorithm does not yield well defined clusters. Note that if the environment was bigger, this overlapping problem would arise for a greater number of proto-objects. It must also be noted that our simulation is not sophisticated enough to properly deal with object occlusions in a consistent way, as a 3D simulation taking the perspective of the agent into account would do. Better dealing with these occlusion issues is left for future work as it requires tackling more difficult questions about memory and the perception of space.

\clearpage

\subsubsection{Influence of the probabilities $p_{obj}$, $p_{env}$, and $p_{abs}$}
\label{sec:penv}

We investigate the impact of the probability of displacement of proto-objects , $p_{obj}$, on the result of the clustering. We run the simulation for several values of $p_{obj}$ between $0$ and $1$, and we show the results in \figurename{~\ref{fig:params}} b). The difficulty of proto-objects discovery increases with their probability of movement. This result is expected because the discovery of proto-objects depends on the probabilities of sensorimotor regularities implied by their structure. These regularities vanish when the expected structure cannot be statistically differentiated from randomness, which  happens when the proto-objects never keep the same position between time steps. Intuitively, this means that if the world around us were to change constantly, we would not be able to discover objects.

\begin{figure}[ht!]
    \centering
    \includegraphics[width=0.82\textwidth]{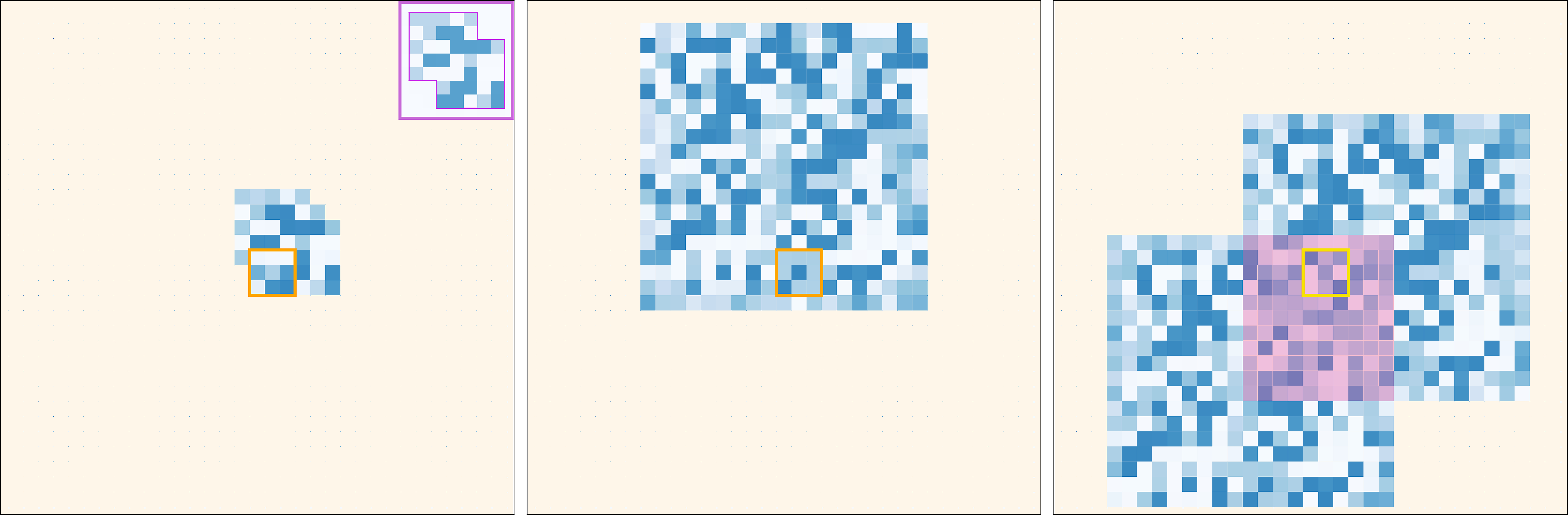}
    \caption[fig]{\textbf{Sensorimotor prediction with a static environment ($p_{env}=0$):} Left: prediction of the pixels corresponding to a proto-object. Center: the entire environment learned by the agent. Right:  given the input state, the prediction associated with some specific movements is ambiguous. The corresponding pixels are highlighted in red. For those, several predictions can contradict each other.}
    \label{fig:pred_stable}
\end{figure}

We also investigate the impact of the probability of updating the environment $p_{env}$ on the result of Spectral Clustering. The simulation is run with $p_{env}$ ranging from $0$ to $1$, with results presented in \figurename~\ref{fig:params} c). When $p_{env}$ is high, for instance when $p_{env}=0.8$, the diagonal of the matrix does not contain high probabilities anymore, since the environment changes too frequently. Although optimal for our simulation, this setup is not realistic considering our own sensorimotor experience, where an environment with no spatio-temporal structure at all is rarely encountered.
Another special case arises when $p_{env}=0$, which means that the environment never changes. Then, the sensorimotor experience while interacting with the environment is completely predictable and the environment should be identified as a third proto-object, as illustrated in the first column of \figurename~\ref{fig:params} d). \figurename{~\ref{fig:pred_stable}} shows sensorimotor predictions for $p_{env}=0$. 
Since the environment never changes, this specific setup highlights sensory ambiguity as one potential limitation of the simulation. Indeed, it is possible for a sensory state to appear in multiple proto-objects, or multiple times in a single object, making it ambiguous. The probability of such a situation is low in the standard setup of the simulation due to the limited size of the proto-objects. However, when $p_{env}= 0$, the whole environment appears as a big proto-object, which significantly increases the probability of encountering ambiguous sensory states. Spectral Clustering is robust to this kind of ambiguity, as it assigns the sensory state to one cluster only, but we can see in the third panel of \figurename{~\ref{fig:pred_stable}} that ambiguity can interfere with sensorimotor prediction. Indeed the reference sensory input seems to appear twice in the constant environment.

As a consequence, the sensory prediction of the agent is a mixture of two contributions that overlap. The pixels which correspond to an ambiguous prediction are highlighted in pink. To disambiguate such a situation, the agent would need to have a memory, or a way to hierarchically extract contexts from its sensorimotor experience, as proposed in \citep{Hemion2017}.

Finally, we analyze the effect of varying the probability of the proto-objects  being absent in the environment. To do so, we run the simulation with changing values of $p_{abs}$ and show the results in \figurename~\ref{fig:params} d). Intuitively, the identification of densely connected subgraphs is easier when the proto-objects  are present at each time step. On the contrary, it becomes harder when $p_{abs}$ is high, since the sensorimotor regularities associated with proto-objects  are encountered with less consistency.

\subsubsection{Rigidly linked proto-objects }

\begin{figure}[t!]
    \centering
    \includegraphics[width=1\textwidth]{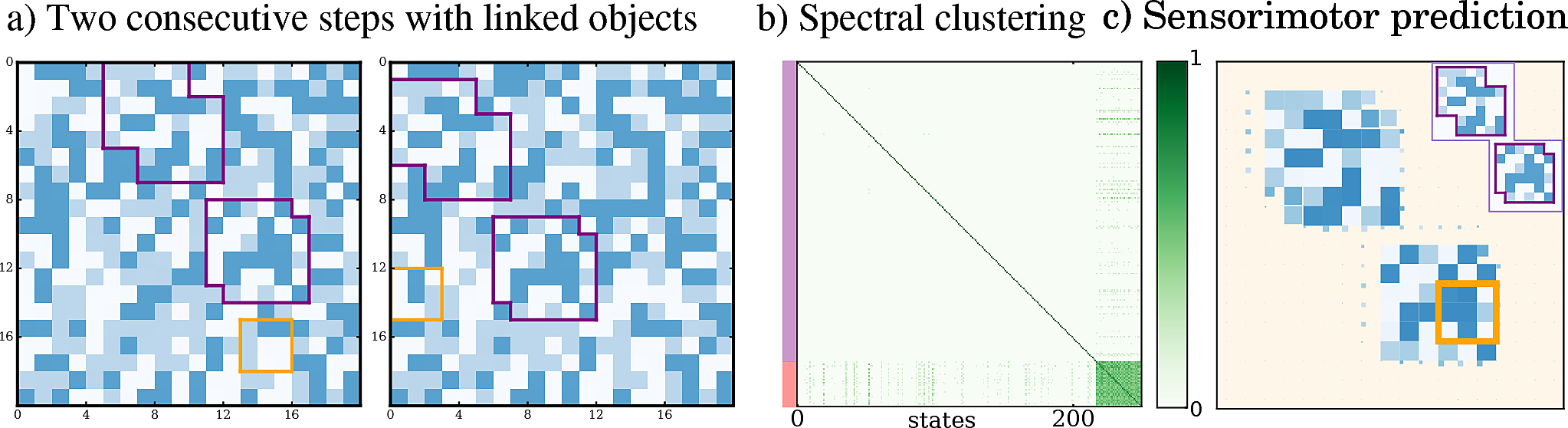}
    \caption[Rigidly linked proto-objects]{\textbf{Rigidly linked proto-objects:}
    \textbf{a)} Two consecutive exploration steps when the proto-objects are linked. Between both steps, the proto-objects keep their relative spatial position.
    \textbf{b)} A single densely connected cluster is identified, corresponding to a single global proto-object identified.
    \textbf{c)} Given an instantaneous sensory input from one proto-object, the agent is able to accurately predict states from the other proto-objects if it performs one of the corresponding movements. 
    }
    \label{fig:split}
\end{figure}

Here we illustrate a property of our definition of proto-objects as spatio-temporally invariant structures. We run a simulation where two proto-objects  are rigidly linked: they move together and thus keep their relative spatial position constant during exploration. In \figurename~\ref{fig:split}, we see that the agent extracts only one densely connected subgraph. This experience of the agent with two linked proto-objects  is thus interpreted as an interaction involving a single proto-object, as the agent extracts a single densely connected graph of sensorimotor transitions. Indeed, the agent looks for sensorimotor regularities without having a notion of spatial contiguity. Thus it does not distinguish the two components of the linked proto-objects . Intuitively this suggests that if we were to live in a world where objects are made of several rigidly linked but disconnected parts, we might interpret them as single entities.

\subsubsection{Identical proto-objects }

\begin{figure}[t!]
    \centering
    \includegraphics[width=0.9\textwidth]{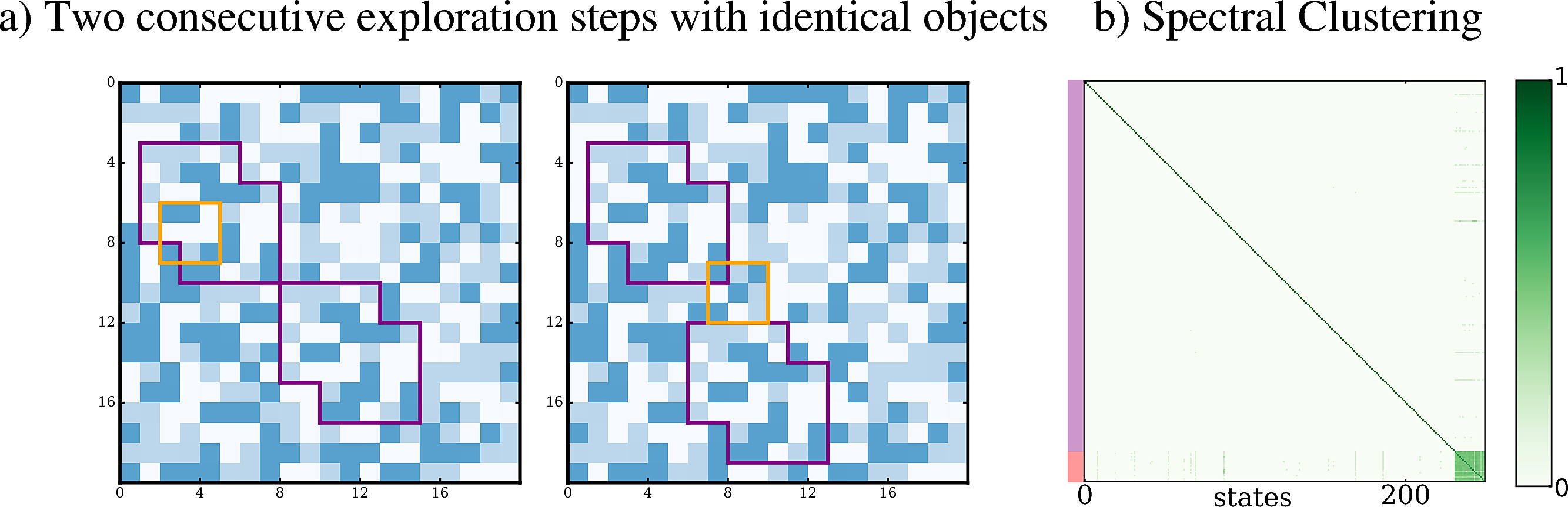}
    \caption[Proto-object with multiple instances]{\textbf{Proto-object with multiple instances:}
    \textbf{a)} Both proto-objects look identical.
    \textbf{b)} A single densely connected subgraph is identified by the Spectral Clustering step.
    }
    \label{fig:id}
\end{figure}

We investigate the special case where both proto-objects  in the environment are identical instances of the same proto-object  . We run the standard simulation where the second proto-objects is a copy of the first one, and show the results in \figurename~\ref{fig:id}. The agent extracts a single densely connected subgraph. This is expected as the agent cannot separate the sensory inputs coming from one instance of the proto-objects from the inputs coming from the other instance. The method can only distinguish types of proto-objects , but not identical instances. A possible solution to separate inputs coming from different instances would be to have a memory and a notion of position in the environment, which the agent does not currently have. 

\subsubsection{Agent rotating the proto-objects }
\label{sec:rotating}

\begin{figure}[t!]
    \centering
    \includegraphics[width=0.9\textwidth]{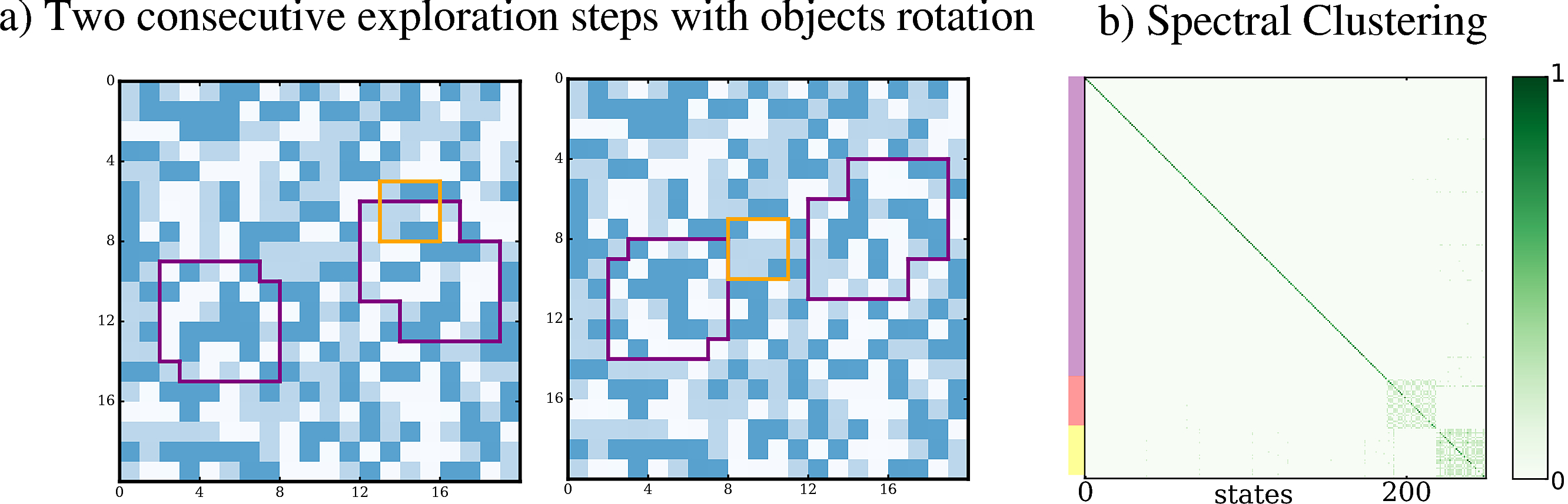}
    \caption[fig]{\textbf{gent rotating proto-objects:} \textbf{a)} Rotating proto-objects generates a different kind of regularity \textbf{b)} The agent is still capable of discovering them.}
    \label{fig:rotation}
\end{figure}

An important aspect of our approach is that the extraction of proto-objects from the environment should not depend on their visual appearance, which means that it does not depend on their pattern of pixels. Additionally, the actions performed by the agent could be of any nature, meaning they are not limited to sensor movements. In order to illustrate these properties, we run a simulation where the agent can move its sensor and also rotate the proto-objects . This action has no effect on the pixels of the environment, but has the consequence of rotating both proto-objects  by 90 degrees . Thus, such a rotation changes the appearance of the proto-objects  and the set of sensory inputs that the agent can receive by interacting with the proto-objects  is larger than when it cannot rotate them. Results of this simulation are presented in \figurename~\ref{fig:rotation}. After exploration and processing of the sensorimotor data, two densely connected subgraphs are still correctly extracted from the experience of the agent. However, it appears that the clusters are slightly less densely connected than in previous simulations. This might come from the \kmeans clustering step, since the sensory inputs are distributed differently in the input space, and from the larger number of possible movements. 

This shows that if the agent performs non-spatial actions, it can still extract structure induced in its sensorimotor flow by the presence of invariant  proto-objects . More generally, any type of action could be performed to learn any structure in the interaction with the world, as long as its effect on the sensory flow of the agent generates some statistical regularities, such as changing the light projected to the global scene, resulting in different pixel values.

\subsubsection{Small proto-object }

\begin{figure}[t!]
    \centering
    \includegraphics[width=0.9\textwidth]{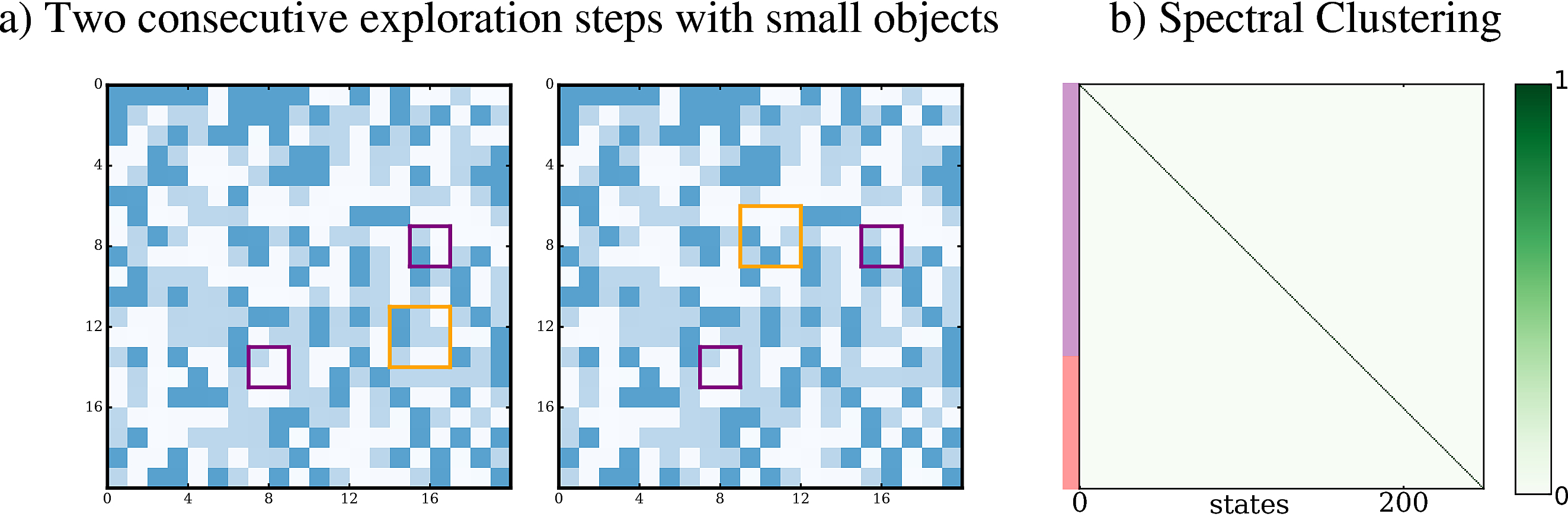}
    \caption[Proto-objects smaller than the receptive field]{\textbf{Proto-objects smaller than the receptive field:} \textbf{a)} The two proto-objects are smaller than the receptive field. \textbf{b)} No densely connected subgraph is identified by the Spectral Clustering.
    }
    \label{fig:small}
\end{figure}

We propose a last simulation in which the proto-objects  are smaller than the receptive field of the agent. Results are shown in \figurename~\ref{fig:small}. No densely connected subgraph is detected by the agent, and it is not able to predict pixels outside the scope of its own receptive field. Since the proto-objects  are smaller than the receptive field, the states obtained after the \kmeans clustering cannot represent the proto-objects  accurately, because they also represent pixels that come from the randomly changing environment. Thus, it is likely that these states mix together sensory inputs coming from proto-objects with sensory inputs coming from the environment. Hence, the sensorimotor structure induced by the presence of proto-objects  in the world is blurred. Thus, proto-objects smaller than the receptive field cannot be discovered by the agent. A possible way to overcome this limitation could be to consider a set of smaller receptive fields and to process them collectively. This is left for future work.

\section{Discussion}	
\label{sec:Discussion}

In this work, we addressed object discovery from a sensorimotor perspective. Taking inspiration from SMCT and predictive coding, we defined proto-objects as spatio-temporally invariant  structures, that an autonomous agent can detect through regularities in its sensorimotor experience when interacting with its environment. More precisely, the agent discovers such proto-objects by collecting sensorimotor transitions and clustering together sensory states according to a sensorimotor similarity, which we derived from a statistical analysis of those transitions. We illustrated the method by applying it to simplistic simulations and outlined some limitations. We now discuss the specificities of our approach with respect to the standard computer vision paradigm and other related work, we highlight some key properties of the model and we point to future work given the limitations we highlighted.

\subsection{Specificities of the paradigm}
\label{sec:paradigm}

In the standard computer vision paradigm, the problem of object identification is generally tackled in a supervised way by training a representation learning algorithm, for instance Deep Convolutional Neural Networks \mbox{
\citep{Gu2015a}}.
These algorithms are trained on a large database of static images containing objects, where the identity of the object is provided as a label (see for instance the well-known ImageNet database \mbox{
\citep{deng2009imagenet}}
). Labelling such databases requires a large human effort which can be mitigated by using semi-supervised or transfer learning approaches, without fundamentally changing the underlying object perception paradigm. In this paradigm, identifying an object consists in extracting from a collection of static images of the same object some invariant  set of visual features which are sufficient for discriminating this object from any other. From an engineering point of view, this paradigm is quite efficient as it provides a working solution for many concrete applications. From a more fundamental standpoint, it captures some important aspects of perception in terms of invariant  visual features which are not captured in our work. But this paradigm goes with some issues, as revealed for instance by failure on adversarial examples \mbox{
\citep{szegedy2013intriguing}}
. Another well-known issue is that relying on external labels makes the agent limited to the recognition of objects present in the database. In that respect, using unsupervised learning methods is mandatory if one wishes to design a truly autonomous learning agent.
Our work reveals a third issue. Indeed, any approach processing static images individually cannot extract any object from our simulations since the distribution of pixel values is the same in proto-objects and in the environment. Thus in our work, we are not interested in the visual features characterizing the appearance of an object, but rather in its spatio-temporal consistency.

Thus our approach is focused on a property of objects that is orthogonal to the one captured by the standard computer vision paradigm. Instead of focusing on the extraction discriminative spatial features in static images, we focus on extracting spatio-temporally invariant  patterns in the sensorimotor flow of the agent.
Our approach has several assets. First, it is unsupervised, as opposed to most approaches to the problem of objects detection and classification outlined above. The agent relies neither on externally provided labels nor on rewards, and does not solve a specific task. It discovers the presence of proto-objects, fundamentally driven by the prediction of its sensorimotor experience, and without knowing the structure of these proto-objects in advance. The agent has prior knowledge neither on its sensory structure, nor on the environment, and not even on the structure of the proto-objects: their number, sizes, shapes, appearances, and positions are unknown.

Importantly, the interaction of the agent with the environment does not have to be spatial: the actions performed do not have to be spatial displacements, such as the translations of the sensor as used in the simulations, and the agent does not need to know its spatial position. More generally, the actions performed by the agent and their effects in the environment can be of any nature, as long as they remain consistent in time. As an example, we have shown in \figurename{~\ref{fig:rotation}} that the agent can extract proto-objects performing actions that modify their appearance by rotating them. 
The determination of the class of actions that are necessary and sufficient to build an artificial object perception system following our approach is an important and open question, left for future work.

\subsection{Related work}
\label{sec:related}

There are other approaches to the problem of artificial perception that exploit either unsupervised learning, the temporal information in the sensory flow of an agent, or the interaction between an agent and its environment.

Unsupervised learning algorithms typically capture statistical structure in the data in order to compress them, hopefully creating more abstract representations 
\citep{Bengio2013}. Despite some interesting attempts around generative models \citep{Doersch2015}, it is still unclear how such statistical method applied to static images could lead to the development of a complete autonomous perceptual system. Most of the time, although pretraining a neural network in an unsupervised way can be used to bootstrap a supervised learning system \citep{Erhan2010}, the representations built this way are interpreted a posteriori by a human.

There are some implementations of unsupervised learning which exploit the temporal link between two successive images. As an example, in \citep{Wang2015}, tracked patches in a video stream are constrained to have similar internal representations. In 
\citep{Torralba}, the built representations are used to predict future states, whereas in \citep{Walker} they are used to define a probability over the trajectories of pixels in an image.

Other approaches to the problem of artificial perception claim that in order to build a truly perceiving agent, it is essential to take its actions into account. Instead of exploiting a mere sensory flow, the actions performed by the agent are processed in parallel. These approaches have been gathered under the term \emph{Interactive Perception} \mbox{\citep{Bohg2016}}. Namely, the actions are used to learn representations consistent with ego-motion in \mbox{\citep{Jayaraman}}, or to predict ego-motion from two successive images in \mbox{\citep{Agrawal2015}}. In \mbox{
\citep{Oh}} representations that allow the prediction of the next image conditioned on the agent's action are learned, while the effect of a physical action on an object is learned in \mbox{\citep{Pinto}}. Both motor and sensory information have also been considered to build state representations consistent with robotic priors in \mbox{\citep{Jonschkowski2015}}.

Our approach is in line with these three paradigms: we process the temporal information between sensory inputs and the interaction of an agent with its environment, through unsupervised learning and a drive for prediction. Compared to the works previously cited, the specificity of ours is that we focus on the identification of spatio-temporally invariant  structures from the sensorimotor flow of the agent.

Learning sensorimotor transition triplets $(\mathbf{s}_t, \mathbf{m}_t, \mathbf{s}_{t+1})$ share some similarity with learning $(object, action,$ $effect)$ triplets in the affordance learning literature \citep{Montesano2008,zech2017computational}, but these triplets are learned based on lower level modules extracting independent visual features for objects, effects, and eventually actions. In that respect, our positioning is more radical than most works in this literature, since we do not call upon such low-level feature extraction.

Other attempts have also been made to propose a computational model allowing for a sensorimotor grounding of knowledge for an artificial agent. One of the closest works with respect to ours in terms of investigating the nature of perception is \citep{hay2018behavior}. In this work, the authors try to demonstrate how a naive agent may extract useful concepts from its sensorimotor experience. However, their concept learning framework assumes that there exists a separate reward function for each concept, an assumption that we consider too strong. In former works investigating the sensorimotor grounding of knowledge, such as \citep{Dorigo1994} and \citep{Scheier}, an external reinforcement signal was also used. In \citep{Cohen1997}, a large amount of semantics is associated a priori  with the actions performed by the agent and with its sensory stimulation, putting this work at a different level of abstraction. Finally, in \citep{Der1999}  an agent uses a model of its sensorimotor interaction with the world in order to optimise at the same time its own structure (the parameters of the body of the agent) and the model itself . However, this work does not propose a mechanism to process the sensorimotor flow of the agent in order to build more abstract knowledge, like our agent does when learning to identify proto-objects as subgraphs in its general sensorimotor experience. In \citep{Maye2011}, an agent learns to predict the effect of its actions on its sensorimotor flow, depending on previous actions and states, learning a model which is very similar to ours. However, while the agent can learn by random exploration, the experimental setup contains no randomness, and the possible actions performed by the agent and its sensory inputs are defined at a more abstract level than ours. Importantly, clustering together sensory states to identify proto-objects is absent from these works, and robustness to randomness in the environment is not studied.

\subsection{Limitations and future work}
\label{sec:related_work}

Despite its versatility, the approach we presented in this paper also suffers from multiple limitations. As revealed in the experiments, our algorithm is not able to handle the case where proto-objects are smaller than the receptive field. In the real world, however, proto-objects appear smaller to us than our field of view. As a consequence, instead of a single receptive field, several elementary receptive fields could be used in combination to define a visual field, as is the case in our own visual system. This should also open possibilities to tackle the problem of distinguishing multiple instances of the same proto-objects, and to reduce the ambiguity of a visual scene. Some preliminary results in this direction have already been published \mbox{
\citep{Laflaqui2016}}
. Instead of considering a small sensor moving in the environment, one could also imagine having a larger sensor with an attention mechanism focusing on a small part of it.

Besides, the implementation presented here was intended to illustrate fundamental mechanisms making it possible to extract proto-objects, but would not scale to a more realistic setting. In a real-life context, the quantification of the sensorimotor experience of the agent would need a way larger amount of memory and computation time, making the method intractable. A more relevant way to process the sensorimotor experience might require an algorithm able to directly process the sensorimotor data without a preliminary \kmeans clustering stage. It should be rather clear from Section~\ref{sec:paradigm} that combining some properties from the standard computer vision paradigm with ours is the way to go in order to address the discovery of objects in real world environments. As an immediate example, using a neural network taking the sensory states and motor commands as input, and predicting the next sensory input could be a promising alternative to the initial \kmeans clustering stage. However, given their very different nature and underlying assumptions, combining both paradigms into a more general framework is a difficult problem which will require careful examination in the future. Finally, learning a more compact representation of the sensorimotor experience, with a tool such as a deep neural network instead of a graph might make it possible to compare our approach to common benchmarks used in the computer vision community.

Finally, when the complexity of the problem increases, or in order to deal with locally ambiguous sensory input, a hierarchical processing of the experience might be necessary. On the one hand, it has been shown that the hierarchical processing of information is probably one of the reasons of the success of deep networks \mbox{
\citep{Lin2017,Bengio2013}}
. On the other hand, from a more biological point of view, it has been shown that biological brains are organized hierarchically \mbox{
\citep{Modha2010}}
, while the interpretation of the reasons for a hierarchical processing have been investigated but are still subject to debate \citep{Damasio1989,Fuster2006}. A proto-object or even an object could then be detected through a hierarchy of features. This approach should also be followed to tackle the problem of ambiguity, by exploiting sequences of transitions in order to define contexts, instead of exclusively exploiting instantaneous transitions \mbox{
\citep{Hemion2017}}
.

\section*{Conflict of Interest Statement}

The authors declare that the research was conducted in the absence of any commercial or financial relationships that could be construed as a potential conflict of interest.

\section*{Author Contributions}

NLH implemented the model, designed and performed the experiments and wrote the paper. AL and OS designed the experiments and wrote the paper.

\section*{Funding}
This work was supported by the Association Nationale Recherche Technologie (ANRT) through a CIFRE convention (contract \textnumero  2016/0946).

\bibliographystyle{unsrt} 
\bibliography{bibliography}


\end{document}